\documentclass[conference]{IEEEtran}
\IEEEoverridecommandlockouts
\usepackage{cite}
\usepackage[labelformat=empty]{caption}
\usepackage{amsmath,amssymb,amsfonts}
\usepackage{algorithmic}
\usepackage{graphicx}
\usepackage{textcomp}
\usepackage{xcolor}
\usepackage{hyperref}
\usepackage{dblfloatfix}
\usepackage{url}
\usepackage{amsmath}
\usepackage{caption}
\usepackage[export]{adjustbox}
\usepackage{amsfonts}
\usepackage{booktabs}
\usepackage{siunitx}

\def\BibTeX{{\rm B\kern-.05em{\sc i\kern-.025em b}\kern-.08em
    T\kern-.1667em\lower.7ex\hbox{E}\kern-.125emX}}
\begin{document}

\title{A Survey on Change Detection Techniques in Document Images}

\author{
\IEEEauthorblockN{Abhinandan Kumar Pun$^{1}$, Mohammed Javed$^{1}$, David S. Doermann$^{2}$}

\IEEEauthorblockA{
\textsuperscript{1}Computer Vision and Biometrics Lab, Department of IT, Indian Institute of Information Technology, Allahabad, India \\
\textsuperscript{2}Department of CSE, University at Buffalo, Buffalo, NY, USA \\
\text{\{mit2021094@iiita.ac.in, javed@iiita.ac.in, doermann@buffalo.edu\}}}
}

\maketitle

\begin{abstract}
The problem of change detection in images finds application in different domains like diagnosis of diseases in the medical field, detecting growth patterns of cities through remote sensing, and finding changes in legal documents and contracts. However, this paper presents a survey on core techniques and rules to detect changes in different versions of a document image. Our discussions on change detection focus on two categories- content-based and layout-based. The content-based techniques intelligently extract and analyze the image contents (text or non-text) to show the possible differences, whereas the layout-based techniques use structural information to predict document changes.
We also summarize the existing datasets and evaluation metrics used in change detection experiments. The shortcomings and challenges the existing methods face are reported, along with some pointers for future research work. 
\end{abstract}

\section{\textbf{INTRODUCTION}}
Identifying areas of change in images/video-frames of the same scene captured at different time intervals is a fascinating research problem in computer vision. According to Radke et al., \cite{radke_image_2005}, the objective of any change detection algorithm is to find the set of pixels that have undergone a significant change between the final image in the sequence and the first image. Some significant and diverse domains where change detection algorithms can be used are remote sensing, assistance for drivers or autopilot systems, diagnosis of diseases and medical treatment, sensing aquatic resources, and video surveillance. In the literature, the researchers have paid much attention to video surveillance and remote sensing. Despite having so many potential applications of image change detection in different areas, a closely related domain like the one to which this survey is dedicated, i.e., change detection in document images, has received significantly less attention.

\begin{figure}[ht]
  \centering
  \includegraphics[page=1,width=.48\textwidth]{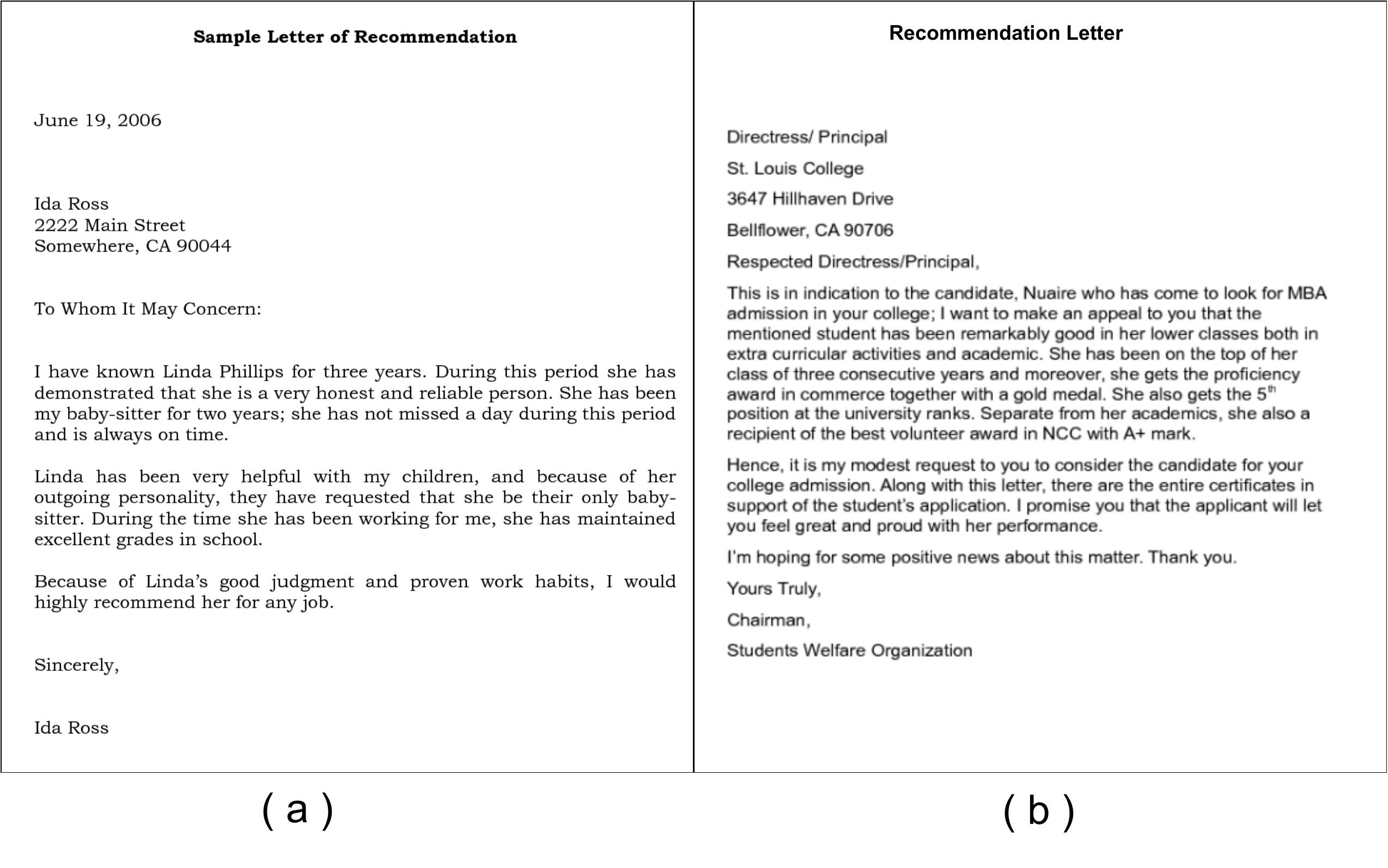}
  \caption{Fig. 1: Two sample document images (a) and (b) having the same layout but with different text contents  }
  \label{fig1}
\end{figure}

\begin{figure}[ht]
  \centering
  \includegraphics[page=1,width=.48\textwidth]{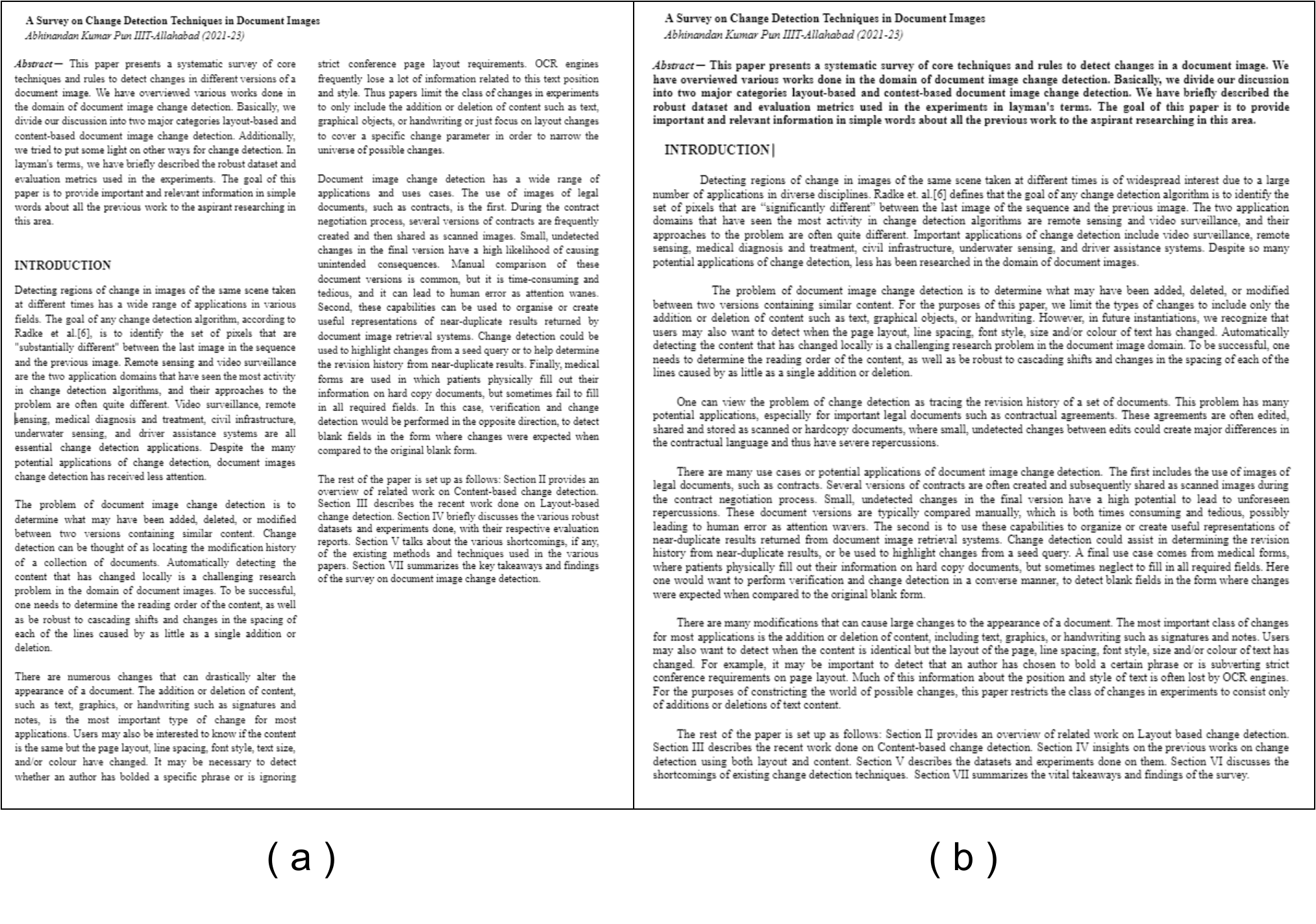}
  \caption{Fig. 2: Two sample document images having the same contents with different layouts shown in (a) two columns and (b) one column }
  \label{fig2}
\end{figure}

Concerning document images, change detection is needed to find the areas within the document where changes have occurred between the two versions of the document. The goal is to identify what may have been added, removed, or updated between the two images with similar contents. In Fig.\ref{fig1}, we have shown two document images having different contents but with the same layout/structure. Moreover, changes are not limited to contents; the same contents can often be arranged with different structures or layouts, as shown in Fig. \ref{fig2}. Change detection can also be considered to locate the modification history of a collection of documents. In document imaging, one challenging task is automatically detecting locally modified contents. Also, the issues like content's reading order, cascading shifts, changes in line spacing, font sizes, and font styles should be intelligently tackled.

Numerous changes can drastically alter the look and even the aspect of a document. The most significant change that applications focus on is content alteration. Inclusion or removal of content, such as a character, word, text, pictures, or pen strokes like handwritten signatures, annotations, and notes. One may also be intrigued to know if the content is the same and if other parameters like the page layout, line spacing, font style, text size, or color have changed. In some situations, it might also be necessary to determine whether a writer bolded a particular word or disregarded the strict guidelines for conference or symposium page layout. Due to these many parameters for change detection, most papers confine the class of changes in experiments to include content or layout changes or only cover a specific change parameter to narrow down the universe of possible changes to consider.

Jain et al. \cite{jain_localized_2015} discusses the wide range of applications and use cases of Document image change detection. Images of legal documents like agreements, bonds, and contracts are widely used during legal proceedings. During the contract negotiation process between two parties, various editions of a contract are frequently revised and shared as digital document images. A minute, undetected change in the final versions of the contract can lead to unintended consequences. Manual comparison of these document versions is standard, but it is time-consuming and tedious and can lead to human error due to attention fatigue. It also plays a significant role in organizing or sorting tidy representations of nearly identical results provided by image retrieval systems. Processing medical forms on which patients fill out their details using scanned image change detection is another application. There might be cases where the patient may fail to fill in all the mandatory fields. In these situations, change detection methods can be applied in reverse by comparing with the initially empty form  to identify and locate empty fields in the form that should have been filled in with the necessary information.

The following paper is structured as follows: Section II provides an overview of related work on Content-based change detection. Section III describes the recent work done on Layout-based change detection. Section IV briefly discusses the various robust data sets and experiments with their respective evaluation reports. Section V discusses the shortcomings of the existing methods and techniques used in the various papers. Section VII summarizes the key takeaways and findings of the survey.

\section{\textbf{CONTENT-BASED CHANGE DETECTION}}
Document change detection is similar to early identical document detection research, which was initially looked into to lower the repetition or redundancy of duplicate documents in sizable databases. It's vital to distinguish identical document detection from near-identical document detection algorithms, commonly used for image retrieval systems but ineffective at determining whether two images are identical. The most straightforward technique for detecting changes researchers use as their first pass is to compare images and identify areas of change that differ by a few pixels. But this method is bounded by many properties of the images, such as the dimensions of the two images must be the same, have the same color scale, etc. While using OCR engines to recognize characters to identify changes, it may most frequently lose a lot of information about this text's position and style. Researchers mostly used feature-based methods over the latter pixel-based approach because it is not invariant skew, scale, rotation, or even intensity, which changes frequently.

\begin{figure}[ht]
  \centering
  \includegraphics[page=1,width=.40\textwidth]{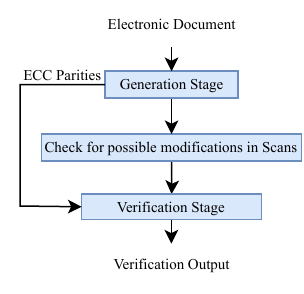}
  \caption{Fig. 3: The working of CIPDEC model \cite{sankarasubramaniam_detecting_2010}}
  \label{fig3}
\end{figure}

Sankarasubramaniam et al. \cite{sankarasubramaniam_detecting_2010}
present the Content Integrity of Printed Documents using Error Correction (CIPDEC) algorithm that detects changes in the printed document. To precisely detect the addition or deletion of any number of pixels, CIPDEC employs an Error Correcting Code approach. The process is divided into two stages: generation and verification. In contrast to the verification stage, which guarantees the authenticity of a printed and scanned document that has been fraudulently altered, generation occurs before a paper document is printed. Any pixel-level change to the printed document can be detected during verification. As shown in Fig. \ref{fig3}, the ECC parities are critical for detecting pixel-level tampering. The ECC parities are computed during the generation phase and used during the verification phase. Any modification done on the paper document is regarded as a pixel error, which the ECC corrects.

Jain et al. \cite{jain_visualdiff_2013}  discussed document change detection and verification. According to their research, document verification should determine if two documents are identical in content, even with pixel value and dimension differences. When document verification fails, change detection comes to the rescue and helps a user to identify the exact difference between two documents. They claim that, to the best of their knowledge, the only way for users to get document images back then was to OCR the documents and compare the resultant text using tools like UNIX diff. A conventional method like the longest common subsequence used to find the text differences was applied to document images. To simulate line-level change detection such as UNIX diff, lines from the document image were first extracted using OCROpus. OCROpus preprocesses by removing skew and noise from the page, then extracts regions from the page layout, extracts lines from each area, and adjusts the baselines of each line before outputting a binarized image of each line on a page. The image segmentation in OCROpus is highly dependent on benchmarks derived from the image resolution. Because text extraction using OCR engines like Tesseract and OCROpus relies highly on image variations, an image-based SIFT feature method was used to extract text from images. Once again, the image-based SIFT features surpassed OCR in performance because OCR can't handle blurry, degraded images or character errors that occur on every line.

\begin{figure}[ht]
  \centering
  \includegraphics[page=1,width=.48\textwidth]{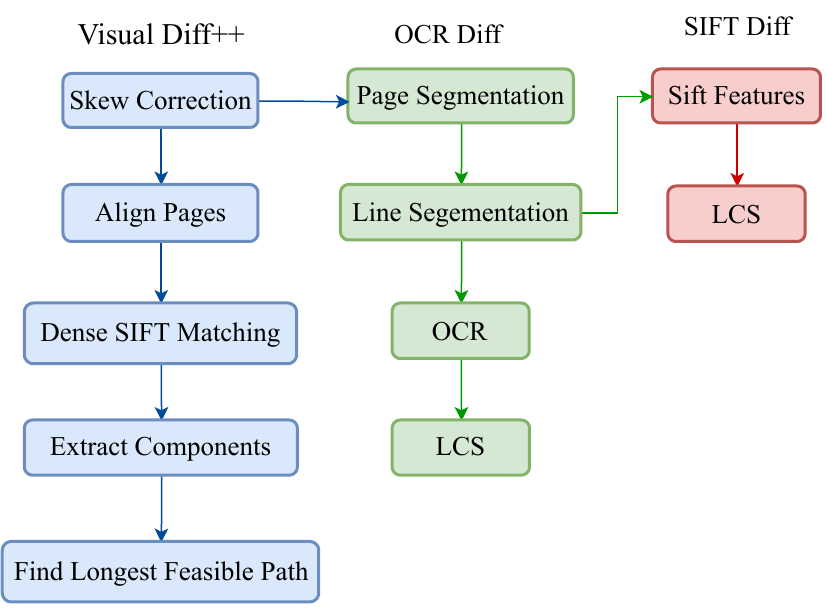}
  \caption{ Fig. 4: The flow diagram for the three techniques- VisualDiff, OCR+LCS, and SIFT+LCS proposed in \cite{jain_localized_2015}}
  \label{fig4}
\end{figure}

Three different methodologies were investigated by Jain et al. \cite{jain_localized_2015}, shown in Fig. \ref{fig4}, for recognizing changes in document images. All these methods aim to minimize false positives while detecting content changes. The longest common subsequence (LCS) algorithm is a reference to find changes in the text extracted using OCRs from the document images. To handle situations where OCR may fail, the second approach builds on their previous work \cite{jain_visualdiff_2013} by using the LCS to perform a "diff" on SIFT features of the line images. The first two methods use on-page and line segmentation to identify the changes, whereas the third technique uses a segmentation-free alignment of SIFT features on the page to do the same. The third technique was better than the other methods because the algorithm was robust for more complex layouts even when the OCR engine could not correctly process non-printed text information like handwritten annotations or signatures or segmentation failed to correctly find all changed text regions on the page.

\section{\textbf{LAYOUT-BASED CHANGE DETECTION}}
The problem of detecting changes in a document's layout attempts to identify differences that occur as new content is added to later document versions. Few papers stress the importance of the layout in analyzing and comprehending the characteristics of a document image. Weihong M. et al. \cite{ma_joint_2020} propose a trainable end-to-end framework for restoring historical document content in the correct reading order. Behind the feature extraction network, two branches, the character and layout branches, are added to this framework. A binary mask is produced by the layout branch, which is based on a fully convolutional network. The binary mask is then subjected to the Hough transform for line detection, and the character results are combined with the layout information to restore document content. To model and understand the interactions between text and layout information across scanned document images, Yiheng X. et al. \cite{xu_layoutlm_2020} propose the LayoutLM framework. This type of work helps document tasks like information extraction from scanned documents.

\begin{figure}[ht]
  \centering
  \includegraphics[page=1,width=.44\textwidth]{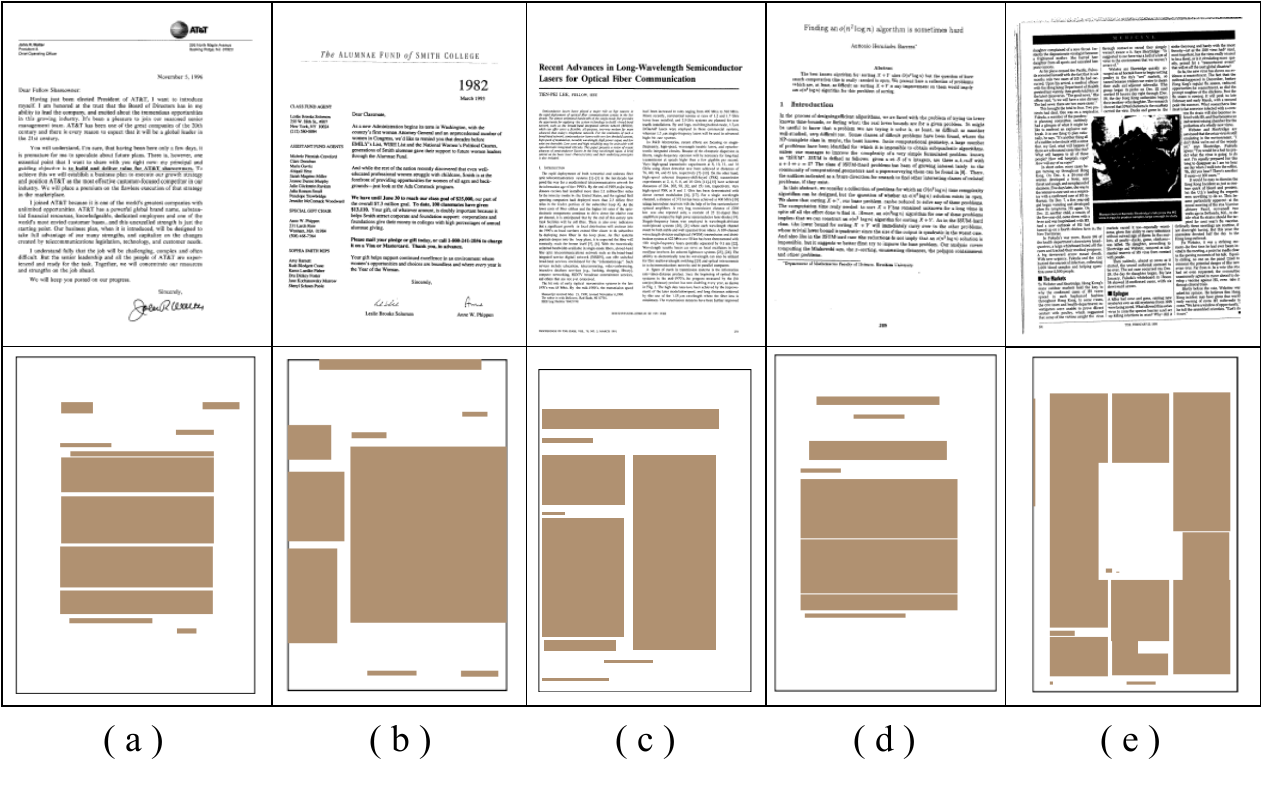}
  \caption{Fig. 5: Different types of documents images with their corresponding block segmentation/layout}
  \label{fig5}
\end{figure}

A lot of work is being done in the field of understanding and analyzing document images through the layout. Compared to the research on content-based change detection, there has been very little research in the domain of change detection in document images. In one of their works, Jianying Hu et al. \cite{hu_document_1999} discuss features and procedures for classifying and comparing document imagery at the spatial layout. In general, measuring spatial layout similarity is problematic because it requires the characterization of similar shapes while accounting for variations arising from design and low-level segmentation process imprecision. Document layout similarity was viewed as a particular case of regional similarity in general images by Zhu and Syeda-Mahmood, who proposed the constrained affine shape model \cite{zhu_image_1998}, a region topology-based shape formalism.

Document Layout Analysis in \cite{oliveira_fast_2017} was done using Fast CNN-based models, and a novel one-dimensional CNN approach was proposed to detect the structure of the documents by recognizing content blocks using horizontal and vertical representations of image tiles. They also implemented a bi-dimensional approach to achieve the same task to compare the results with the first method. Moreover, for document layout  recognition and its analysis, several computer vision algorithms and Deep Learning methods have been developed, such as Layout Parser, Detectron2, OCRFeeder, etc.

\begin{figure*}[htbp]
\centering
  \includegraphics[page=1, width=1\textwidth]{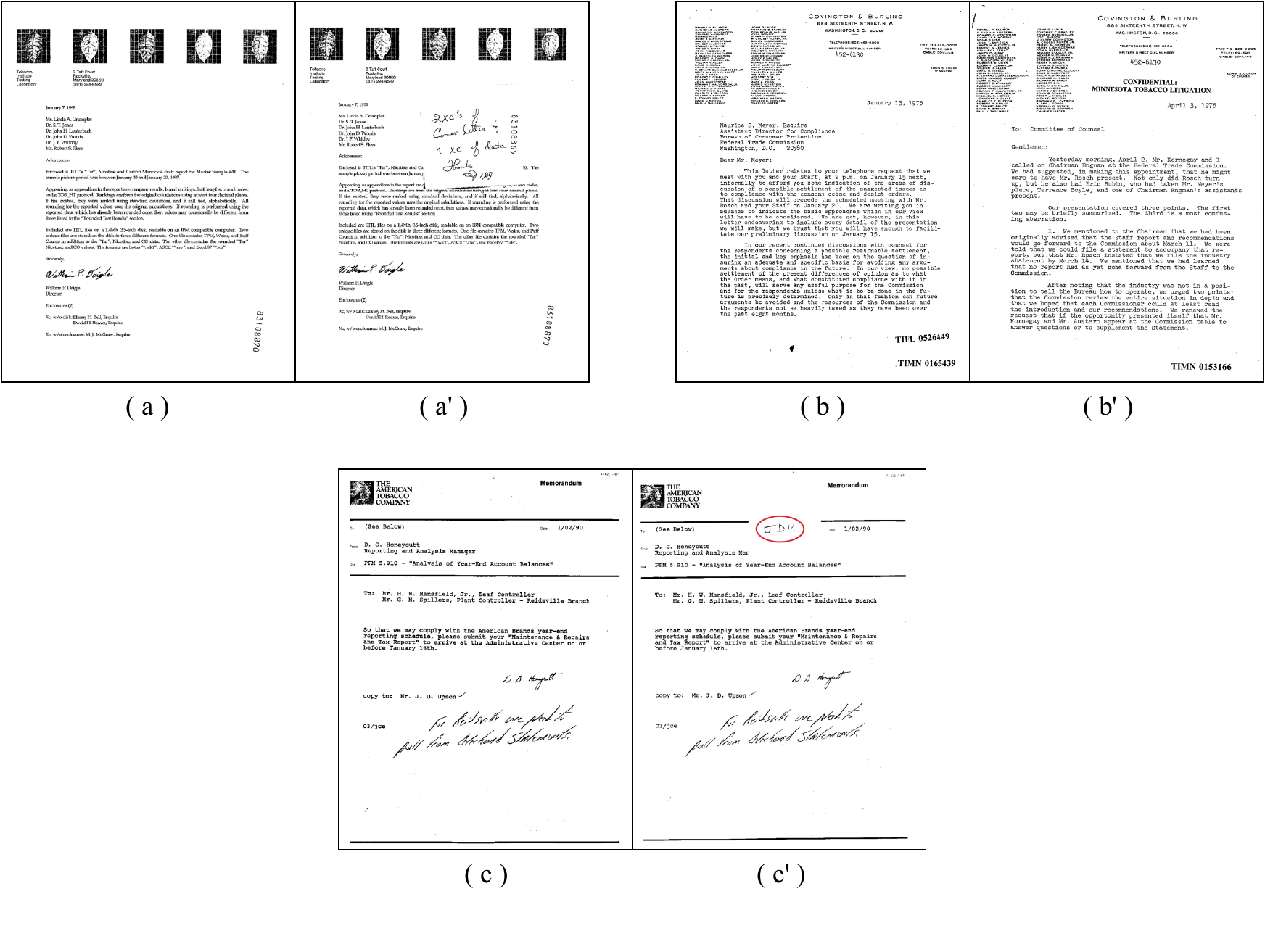}
  \caption{ Fig. 6: Illustrating some challenging cases of change detection from sample images taken from Tobacco Revision Dataset, where (a), (b) and (c) are original documents, and (a'), (b') and (c') include added/modified/deleted contents   }
  \label{fig6}
\end{figure*}

\section{\textbf{DISCUSSION ON DATASETS AND EXPERIMENTS}}
The challenge of document change detection can be researched by following its updated history. To create a ground truth dataset for the experiment on the change detection methods, snapshots of a document's before and after can be obtained from a set of revisions or versions. 

The Enron document collection \cite{klimt_enron_2004} was used for experiments in \cite{jain_visualdiff_2013}. The dataset stores nearly 500,000 files, including over 180,000 samples of Microsoft Word documents. From this collection of documents, 150 one-page documents were selected, which have 4-41 additions or deletions of text. There could be one character to an entire paragraph modification in the selected documents. A hard copy of these before and after images were printed. For the study of document verification and change detection in \cite{jain_visualdiff_2013}, 11 images were captured for 300 pages, i.e., both sides of the mentioned 150 pages. Six variations were created using different DPI settings from 100, 200, and 300 DPI with binarized and non-binarized scans. Then using an iPad cam scanner, the last five images were produced, applying binarization and different blur levels on the original image. This resultant dataset was used to evaluate the methods described in \cite{jain_visualdiff_2013}. Table \ref{tab1}. summarizes the outcomes of the experiments in terms of accuracy, precision, and recall. Data in rows-1 and columns-1,2 shows results for the two methods based on OCR, which is used for character recognition, then applying string matching algorithms like LCS and edit distance on the characters. The last column in rows-1 contains the result of the SIFT-based approach that is mentioned in section II. 


According to Jain et al., \cite{jain_localized_2015}, the Enron Revision dataset works well for studying document change detection in controlled conditions with various variations and constraints. of images in play, it may fall short of accurately capturing the complexities of real-world document image collections. These collections might have more factors, noise, and visually appealing page designs. To solve this, a new Dataset identified nearly identical content changes between two tobacco collection documents, including binarized document images imaged at resolutions of 100 to 300 DPI. The first 100 pairs of nearly identical document images were retained, while actual replicas and completely irrelevant document images were discarded. These documents presented a more significant challenge because they comprised extra information in handwritten notes, signatures,  tables, logos, stamps, graphics, and various layouts, as seen in Fig. \ref{fig6}. The three images a, b, and c, show the document's original version, with the modified document image alongside its respective original, namely, a$'$, b$'$, and b$'$. One can notice each word, signature, and graphic that was annotated are either the same in both versions of the document or were altered.

\begin{table*}[t]
  \small
  \centering
\caption{ Table I: Summary of the existing state-of-the-art work on the problem of content change detection}
    \begin{tabular}{ c c c c c}
    \hline\\
    \textbf{Paper} & \textbf{Evaluation Metric} & &\textbf{Methods} & \\
    \hline
    & & OCR+LCS & OCR+LCS+Edit Distance & SIFT+LCS \\
    \midrule
    & TPR & 90\% & 90\%  & 90\% \\
    Rajiv et al.\cite{jain_visualdiff_2013}& FPR & 19.50\% & 19.60\% & 9.50\%\\
    & AUC & 0.861 & 0.833 & 0.921\\
    & EER & 0.147 & 0.153 & 0.1\\
    \hline
    & & OCR Diff & SIFT Diff & VisualDiff++\\
    \hline
    & Recall & 99.30\% & 96.70\% & 98.20\% \\
    Rajiv et al.\cite{jain_localized_2015}& Accuracy & 49.80\% & 70.90\% & 91.10\%\\

    & Precision & 36.20\% & 49.50\% & 77.00\% \\
    \hline
    \end{tabular}
  \label{tab1}
\end{table*}

Each of the three approaches, namely OCR diff, SIFT diff, and VisualDiff++, was assessed on the 100 pages from the Tobacco NearDupe Dataset. In \cite{jain_localized_2015}, VisualDiff++  outscored the prior techniques, decreasing the False Positive Rates by 50\% and Equal Error Rate by 32\%. This was mainly due to the method's capability to deal with many complex layouts when segmentation ceased to accurately locate all of the regions of interest on the page or when the OCR engine could not recognize pen strokes like handwritten notes, signatures, or annotations. Table \ref{tab1} shows the evaluation metrics for the three approaches: TPR (True Positive Rate), AUC(Area under ROC curve), FPR, and EER.  Rajiv et al.\cite{jain_visualdiff_2013} The area under the curve(AUC) and equal error rate(EER) are reported along with the true positive rate(TPR) and fals positiv rate(FPR) at the point on the operating curve with the highest recall for the baseline OCR diff approach.

\begin{equation}
    \small
     Precision = \frac{t_{p}}{t_{p}+f_{p}}
\end{equation}
\begin{equation}
    \small
     Recall = \frac{t_{p}}{t_{p}+t_{n}}
\end{equation}
\begin{equation}
    \small
    Accuracy =\frac{T_{p}+T_{n}}{T_{p}+F_{p}+T_{n}+F_{n}}
\end{equation}
  
 To justify the purpose of this paper, we have briefly described, without going into much detail, the various methods and their respective evaluation metrics sourced from the papers on content-based document image change detection.

\begin{figure}[ht]
  \centering
  \includegraphics[page=1,width=.4\textwidth]{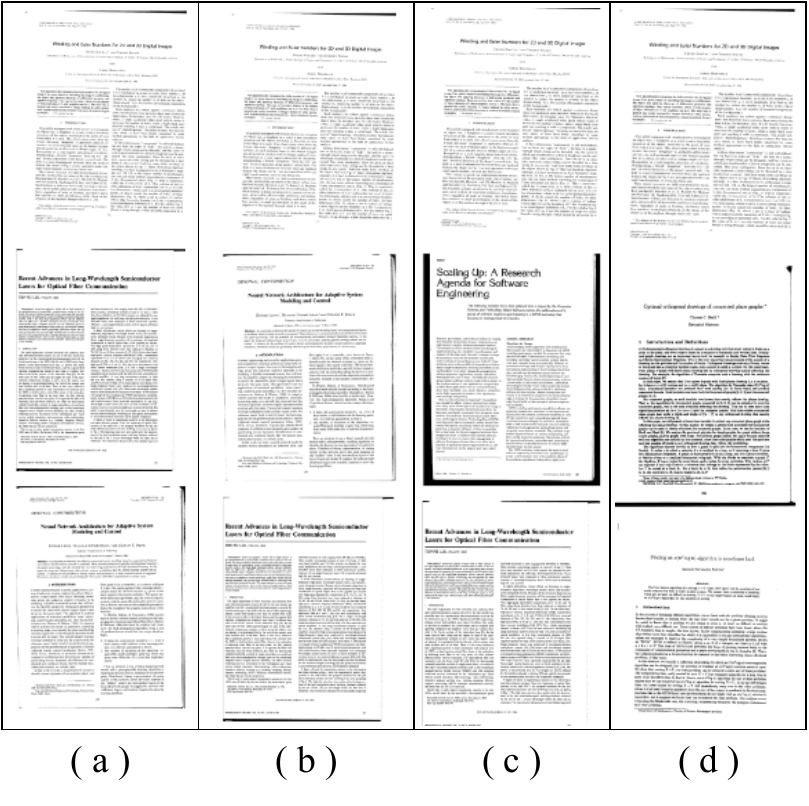}
  \caption{Fig. 7:  Page layout similarity ranking using different distance measures. (Source: Jianying Hu, et. al.\cite{hu_document_1999})}
  \label{fig7}
\end{figure} 

Experiments in \cite{hu_document_1999} were conducted by taking samples of one-column letters, two-column journal pages, two-column journal pages, and magazine articles. A two-level process was adopted to compare the two pages layout. For the first level, they discussed determining the distance between each row of two pages. For the second level locating a connection between the line of the two pages is required to minimize the total distances between corresponding rows. This is achieved using dynamic programming paradigm \cite{sakoe_dynamic_1978}, which helps find optimal paths by aligning the rows on two pages while minimizing the total distance over a finite range of vertical shifts. Four distance measures were used here: edit distance, cluster distance, bitmap distance, and interval distance. The objective of the experiment on layout comparison was to order a collection of documents in ranks in terms of their likeness to a test document; ten samples from each of the five types were used. After that, a document was chosen spontaneously as a test sample, and ordering was done for the left documents in terms of their likeness to the test document.

In Fig.\ref{fig7}, the experiment results can be seen; the three most similar samples are listed to a standard original image using the three different distance measures. Images a, b, c, and d show the images that were found similar using edit distance, interval distance, cluster distance, and bitmap distance, respectively.

\section{\textbf{SHORTCOMINGS IN EXISTING CHANGE DETECTION TECHNIQUES}}
When OCR was first used in document change detection in both \cite{jain_visualdiff_2013} and \cite{jain_localized_2015}, it encountered a few issues. A good preprocessing to capture good lines and page segmentation is fundamental for the methodology to be more precise and accurate. Otherwise, poor segmentation of an image could lead to large document segments being evaluated for changes. The Open Source OCR services showed poor results for the Enron Revisions dataset. Even while using a commercial OCR service, there was a word error. In cases of images taken from mobile, the Error rate was more than 70\% as the samples were blurred and low in quality.

Secondly, the error rate in OCR can differ considerably from commonly researched, simplified scripts like Latin to more complex languages like Mandarin and Devanagari. It can frequently result in false positives. Finally, because OCR engines are designed to work on machine print text, they may not process pictures, artworks, logos, trademarks, emblems, stamps, handwritten corrections, annotations, signatures, and much more essential information, thus missing significant changes.

In \cite{jain_localized_2015}, the SIFT-based method also suffers from segmentation error as when text is added or removed, it may cause the shift of the line to the right, or it can even have a domino effect on the consecutive lines of a paragraph making it shift directly down. For different layouts, there might be different shifts; for example, given a 2-column page, a case may exist where adding some content may shift the following content in the paragraph from the bottom left of the first column to the upper right of the second column. In such cases, SIFT-based techniques fail to be segmentation free.

\section{\textbf{CONCLUSION}}
Apart from content-based and layout-based change detection in document images, \cite{kim_detection_2018} presents models based on deep neural networks that focus on detecting the semantic and structural modifications in text documents.  As Deep learning has seen significant interest over the past decade for its state-of-the-art algorithms and performance, researchers can harness its robust capabilities to solve the problem of document image change detection. So, having discussed the potential applications and different techniques of document change, we can conclude that the domain of detecting changes in the document image, to the best of our knowledge, is relatively unexplored. On the contrary, there are pretty potential applications of change detection. All of the approaches discussed in this survey could use some work. Discovering additional types of changes, such as substitutions, formatting, and change in font, or categorizing the types of changes present in documents, such as font size change, font style change, text replacement, addition, or deletion, are possible extensions of the work.

\bibliographystyle{plain}
\bibliography{reference}

\begin{thebibliography}{10}

\bibitem{hu_document_1999}
Jianying Hu, R.~Kashi, and G.~Wilfong.
\newblock Document image layout comparison and classification.
\newblock In {\em Proceedings of the {Fifth} {International} {Conference} on
  {Document} {Analysis} and {Recognition}. {ICDAR} '99 ({Cat}.
  {No}.{PR00318})}, pages 285--288, September 1999.

\bibitem{jain_visualdiff_2013}
Rajiv Jain and David Doermann.
\newblock {VisualDiff}: {Document} {Image} {Verification} and {Change}
  {Detection}.
\newblock In {\em 2013 12th {International} {Conference} on {Document}
  {Analysis} and {Recognition}}, pages 40--44, August 2013.
\newblock ISSN: 2379-2140.

\bibitem{jain_localized_2015}
Rajiv Jain and David Doermann.
\newblock Localized document image change detection.
\newblock In {\em 2015 13th {International} {Conference} on {Document}
  {Analysis} and {Recognition} ({ICDAR})}, pages 786--790, August 2015.

\bibitem{kim_detection_2018}
Noo-ri Kim, YunSeok Choi, HyunSoo Lee, Jae-Young Choi, Suntae Kim, Jeong-Ah
  Kim, Youngwha Cho, and Jee-Hyong Lee.
\newblock Detection of document modification based on deep neural networks.
\newblock {\em Journal of Ambient Intelligence and Humanized Computing},
  9(4):1089--1096, August 2018.

\bibitem{klimt_enron_2004}
Bryan Klimt and Yiming Yang.
\newblock The {Enron} {Corpus}: {A} {New} {Dataset} for {Email}
  {Classification} {Research}.
\newblock In Jean-François Boulicaut, Floriana Esposito, Fosca Giannotti, and
  Dino Pedreschi, editors, {\em Machine {Learning}: {ECML} 2004}, Lecture
  {Notes} in {Computer} {Science}, pages 217--226, Berlin, Heidelberg, 2004.
  Springer.

\bibitem{ma_joint_2020}
Weihong Ma, Hesuo Zhang, Lianwen Jin, Sihang Wu, Jiapeng Wang, and Yongpan
  Wang.
\newblock Joint {Layout} {Analysis}, {Character} {Detection} and {Recognition}
  for {Historical} {Document} {Digitization}.
\newblock In {\em 2020 17th {International} {Conference} on {Frontiers} in
  {Handwriting} {Recognition} ({ICFHR})}, pages 31--36, September 2020.

\bibitem{oliveira_fast_2017}
Dário Oliveira and Matheus Viana.
\newblock {\em Fast {CNN}-{Based} {Document} {Layout} {Analysis}}.
\newblock October 2017.
\newblock Pages: 1180.

\bibitem{radke_image_2005}
R.J. Radke, S.~Andra, O.~Al-Kofahi, and B.~Roysam.
\newblock Image change detection algorithms: a systematic survey.
\newblock {\em IEEE Transactions on Image Processing}, 14(3):294--307, March
  2005.
\newblock Conference Name: IEEE Transactions on Image Processing.

\bibitem{sakoe_dynamic_1978}
H.~Sakoe and S.~Chiba.
\newblock Dynamic programming algorithm optimization for spoken word
  recognition.
\newblock {\em IEEE Transactions on Acoustics, Speech, and Signal Processing},
  26(1):43--49, February 1978.
\newblock Conference Name: IEEE Transactions on Acoustics, Speech, and Signal
  Processing.

\bibitem{sankarasubramaniam_detecting_2010}
Yogesh Sankarasubramaniam, Badri Narayanan, Kapali Viswanathan, and Anjaneyulu
  Kuchibhotla.
\newblock {\em Detecting {Modifications} in {Paper} {Documents}: {A} {Coding}
  {Approach}}, volume 7534.
\newblock January 2010.
\newblock Pages: 10.

\bibitem{xu_layoutlm_2020}
Yiheng Xu, Minghao Li, Lei Cui, Shaohan Huang, Furu Wei, and Ming Zhou.
\newblock {LayoutLM}: {Pre}-training of {Text} and {Layout} for {Document}
  {Image} {Understanding}.
\newblock In {\em Proceedings of the 26th {ACM} {SIGKDD} {International}
  {Conference} on {Knowledge} {Discovery} \& {Data} {Mining}}, {KDD} '20, pages
  1192--1200, New York, NY, USA, August 2020. Association for Computing
  Machinery.

\bibitem{zhu_image_1998}
Wei Zhu and T.~Syeda-Mahmood.
\newblock Image organization and retrieval using a flexible shape model.
\newblock In {\em Proceedings 1998 {IEEE} {International} {Workshop} on
  {Content}-{Based} {Access} of {Image} and {Video} {Database}}, pages 31--40,
  January 1998.

\end{thebibliography}

\end{document}